\documentclass[conference]{IEEEtran}
\IEEEoverridecommandlockouts
\usepackage{cite}
\usepackage{amsmath,amssymb,amsfonts}
\usepackage{algorithmic}
\usepackage{graphicx}
\usepackage{url}
\usepackage{textcomp}
\usepackage{xcolor}
\usepackage{subcaption}
\def\BibTeX{{\rm B\kern-.05em{\sc i\kern-.025em b}\kern-.08em
    T\kern-.1667em\lower.7ex\hbox{E}\kern-.125emX}}

\makeatletter 
\newcommand{\linebreakand}{%
  \end{@IEEEauthorhalign}
  \hfill\mbox{}\par
  \mbox{}\hfill\begin{@IEEEauthorhalign}
}
\makeatother 

\begin{document}

\title{TSLFormer: A Lightweight Transformer Model for Turkish Sign Language Recognition Using Skeletal Landmarks\\
\thanks{}
}

\author{
    \IEEEauthorblockN{Kutay ERTÜRK}
    \IEEEauthorblockA{\textit{Department of Electrical and Electronics Engineering} \\
    \textit{Eskişehir Technical University}\\
    Eskişehir, Turkey \\
    kutayerturk@ogr.eskisehir.edu.tr\\
    ORCID: 0009-0000-9612-8954}
    \and
    \IEEEauthorblockN{Furkan ALTINIŞIK}
    \IEEEauthorblockA{\textit{Department of Electrical and Electronics Engineering} \\
    \textit{Eskişehir Technical University}\\
    Eskişehir, Turkey \\
    furkanaltinisik@ogr.eskisehir.edu.tr}
    \linebreakand
    \IEEEauthorblockN{İrem SARIALTIN}
    \IEEEauthorblockA{\textit{Department of Electrical and Electronics Engineering} \\
    \textit{Eskişehir Technical University}\\
    Eskişehir, Turkey \\
    iremsarialtin@ogr.eskisehir.edu.tr}
    \and
    \IEEEauthorblockN{Prof. Ömer Nezih GEREK}
    \IEEEauthorblockA{\textit{Department of Electrical and Electronics Engineering} \\
    \textit{Eskişehir Technical University}\\
    Eskişehir, Turkey \\
    ongerek@eskisehir.edu.tr\\
    ORCID: 0000-0001-8183-1356}
}

\maketitle

\begin{abstract}
This study presents TSLFormer, a light and robust word-level Turkish Sign Language (TID) recognition model that treats sign gestures as ordered, string-like language. In contrast to working with raw RGB or depth videos, our method only works with 3D joint positions—articulation points—extracted using Google's Mediapipe library, which focuses on the hand and torso skeletal locations. This creates efficient input dimensionality reduction with significant preservation of important semantic information of the gesture.
Our approach revisits sign language recognition as sequence-to-sequence translation, drawing inspiration from sign languages' linguistic nature and transformer's success at natural language translation. Since TSLFormer adapts the transformers' self-attention mechanism, it effectively represents the temporal co-occurrence of a sign sequence, stressing significant movement habits over time as words are referenced in a sentence.
Experimented and validated on the AUTSL dataset holding over 36,000 sign samples of over 226 different words, the TSLFormer achieves competitive performance and with minimal computational demands. From the experimentation, rich spatiotemporal understanding of signs is evidenced, and using only joint landmarks, it is possible within any real-time, mobile, and assistive technology facilitating communication between hearing-impaired members.
\end{abstract}


\section{Introduction}

Sign language is an essential communication method for the hearing impaired to express ideas and sentiments through hand gestures, facial expressions, and body movement. Unlike spoken languages, which employ auditory and verbal modalities, sign language utilizes visual and spatial modalities to express meaning. However, despite the limited number of sign language proficient individuals, communication gaps still exist to hinder inclusion—particularly in social interaction on a daily basis and in employment, educational, and healthcare environments.

It is against this gap that automatic sign language recognition (SLR) systems have been mooted as a potential solution, which leverages advancements in artificial intelligence, computer vision, and deep learning to interpret visual gestures and translate them into text or speech. While there has been significant work on classifying static gestures, achieving high accuracy on real-world, dynamic sign recognition remains challenging. Most models are not yet capable of capturing the rich temporal dynamics of sign language, especially when gestures vary in speed, intensity, or execution across individuals.

Turkish Sign Language (TİD) has two main systems: an alphabet system in which signs are static handshapes for individual letters, and a word system in which entire words are represented using dynamic, fluid hand and arm movement. Alphabet recognition has been heavily studied, often with static image-based models such as convolutional neural networks (CNNs), but word recognition is a more complex problem including understanding the sequential and temporal character of motion.

For this purpose, we present TSLFormer, a word-level Turkish Sign Language recognition model based on the transformer. Our approach treats sign language not as a simple sequence of visual cues, but as a real language with grammar and sequence context, the same as spoken or written language. In this regard, we draw inspiration from the achievements of large language models (LLMs) for natural language processing (NLP) and apply the same principles to sign language, using the transformer's attention mechanism to capture the temporal dependencies of gesture sequences. This framing allows the model to read a sign gesture the same way it would read a sentence—by attending to the sequence, duration, and significance of each visual "token" over time.

TSLFormer employs Google's Mediapipe framework \cite{lugaresi2019mediapipeframeworkbuildingperception} to obtain 3D skeletal landmark features of 16 evenly sampled frames within a video, but only taking into account salient joint positions of the hands and upper body. This bypasses the need to process raw RGB frames or depth data, thereby conserving computational complexity without compromising on the semantic essence of the gesture.

For training and testing of our system, we use the AUTSL dataset \cite{Sincan_2020}, a large Turkish Sign Language dataset containing over 36,000 video samples for 226 different words signed by 43 different signers. The variation in signers and recording conditions exposes the model to natural variations in gesture execution and enhances its generalization capability.

By combining landmark-based visual abstraction with the transformer’s capacity for sequential modeling, TSLFormer demonstrates a highly efficient and scalable solution for Turkish Sign Language word recognition. Our goal is to contribute to the broader effort of making communication more inclusive by developing lightweight models suitable for real-time applications, mobile deployment, and assistive technologies for the hearing-impaired community.

\section{Related Work}

Sign language recognition has been one of the widely studied fields in computer vision and deep learning, with various approaches focusing on alphabet-based and word-based recognition methods. Many early studies explored static gesture recognition, while more recent work expanded into dynamic gesture classification using recent deep learning techniques emerged. The effectiveness largely depends on the feature extraction techniques and the chosen model architecture, especially when dealing with video sequences and sequential motion patterns.

Hidden Markov Models (HMMs) are one of the first methods used in sign language recognition \cite{HMMS}. HMMs used to be widely applied in speech and gesture recognition because they can model sequential data over time. They can recognize hand movements and transitions between the gestures in sign language. Starner and Pentland developed a system which recognizes American Sign Language (ASL), based on HMMs, which could recognize full sentences using only visual inputs \cite{asl}. Similarly, Yang et al. proposed an HMM-based system for Chinese Sign Language that combined hand trajectory and shape features, proving that HMMs could be used across different sign languages \cite{chinese}. However, while HMMs were useful in earlier studies, they have several limitations. They require extracting the features from videos manually, meaning that researchers must underline the important movement patterns. If the researchers don't mark the important features, HMMs cannot work effectively. Also, if we work with large and diverse datasets, HMMs cannot generalize well. As a result, deep learning models are usually preferred more nowadays, because they can learn features directly from raw images and videos, achieving better performance.

With the development of deep learning, Convolutional Neural Networks (CNNs) became one of the most popular approaches for static hand gesture recognition \cite{oshea2015introductionconvolutionalneuralnetworks}. CNNs are great at image classification, making them ideal for recognizing individual hand shapes in alphabet-based sign language. The most popular CNN architectures for the static gesture recognition have been architectures like LeNet, VGGNet, and ResNet \cite{simonyan2015deepconvolutionalnetworkslargescale, lenet, he2015deepresiduallearningimage}. When these models are trained on well-labeled datasets ,these models can achieve high accuracy. Pigou et al. applied deep CNNs to recognize the fingerspelling gestures in ASL, and the results have been promising \cite{cnn}. Buckley et al. trained a CNN with a British Sign Language dataset, and they received an 89\% accuracy with still images \cite{britcnn}. However, while CNNs can be applicable for single-frame classification, they cannot be effective with continuous sign gestures, making them less suitable for recognizing full words or phrases from video sequences.

To handle dynamic gestures, researchers added a temporal dimension to CNNs, and invented 3D Convolutional Neural Networks (3D CNNs). These models can also capture movement patterns across video frames, on top of regular CNNs. Unlike standard CNNs, which process each frame separately, 3D CNNs analyze motion across multiple frames, making them well-suited for video-based sign language recognition. For example, Tran et al. introduced a new 3D CNN method, C3D for usage in the action recognition fields, showing that it works well for understanding movement in videos \cite{tran2015learningspatiotemporalfeatures3d}. Hara et al. combined 3D CNNs and ResNet architectures to build a model for a useful action recognition \cite{3dres}. However, despite their advantages, 3D CNNs need large-scale labeled datasets and high computational power, making them difficult to use in real-world applications where resources are limited.

Also, traditional machine learning methods like Support Vector Machines (SVMs) were used earlier for sign language recognition. SVMs classify gestures using features extracted from hand contours, movement patterns, and articulation points from the video data. SVMs also do not require massive datasets, unlike deep learning models. For example, Rautaray and Agrawal used SVMs to classify hand shapes and motion features, achieving reasonable accuracy \cite{svm}. However, SVM-based approaches usually struggle with high-dimensional data, making them less effective than modern deep learning techniques when working with complex gestures.

Some ensemble methods which combines these methods is applicable as well. In the field of sign language recognition, SVM and CNN methods are the most common methods to be used together. 

But recently, transformers have become a powerful alternative to the methods listed above. Transformers are excellent at capturing long-range dependencies in data and have been adapted for vision-based applications. Some researchers have explored Vision Transformers (ViTs) and self-attention-based networks for gesture classification shows promising results in recognizing motion patterns over time. Camgoz et al. introduced a Sign Language Transformer (SLT), which can perform both gesture recognition and translation, demonstrating that transformer-based models can improve sign language research. Ghadami et al. also used transformers for recognition of the Iranian Sign Language \cite{ghadami2024transformerbasedmultistreamapproachisolated}. It has resulted in a good output of 90.2\% accuracy. However, transformers require a lot of labeled data, and their computational cost can be a challenge, especially for real-time applications \cite{camgoz}. ViViT (Video Vision Transformers) is a transformer model designed for video processing and has been applied in sign language recognition tasks \cite{arnab2021vivitvideovisiontransformer}. However, since ViViT processes entire video frames—including a large amount of unnecessary visual information, it was not preferred in this study.

Another problem of sign language recognition involves extracting skeletal articulation coordinates, which provide a more structured representation of hand and arm movements. Xbox Kinect, a depth-sensing device that can track hand and body movements in three-dimensional space, is widely used for this aim, despite its main purpose of usage, which is playing video games. Kinect-based systems have been employed in several studies to map sign gestures into coordinate-based features, reducing the reliance on raw video processing and removing the unnecessary features from being processed \cite{Berkan}. While this method allows for more structured motion analysis, its effectiveness depends on Kinect's accuracy and various environmental conditions, as variations in lighting and occlusions can affect gesture tracking.

While these improvements in the field of sign language recognition is effective, it is not enough. There is still a need for models which require less computational power, and work more effectively.

\section{Methodology}

\subsection{AUTSL Dataset}

In our research, the utilized dataset is the standard AUTSL dataset, which is designed specifically for Turkish Sign Language recognition \cite{Sincan_2020}. The extensive dataset includes over 38,000 videos of 226 various words in Turkish sign language. The signs were executed by 43 different individuals and under different conditions so that the model could learn how signs are performed differently by different people.

\begin{figure}[h]
    \centering
    \includegraphics[width=0.4\textwidth]{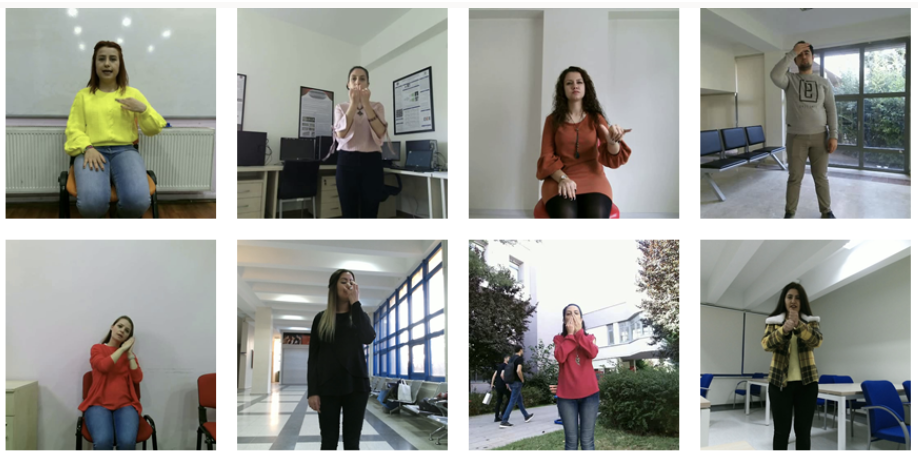}
    \caption{Some Sampled Frames from the AUTSL Dataset}
    \label{fig:example}
\end{figure}

The dataset includes one isolated word in each video of the dataset. The videos were recorded in indoor and outdoor environments, and they represent varied backgrounds and lighting conditions. The subjects in the videos also present different positions and styles, making the dataset more realistic and challenging.

The aim of this dataset is to allow researchers to train and test machine learning models for the recognition of Turkish sign language. In the first study in which AUTSL was introduced, the researchers used deep learning models that first extract features from the video using CNNs, and then used LSTM models to learn the development of the features in time. Furthermore, they used methods like temporal attention to highlight important frames in the video. These models obtained up to 95.95\% accuracy in certain test conditions. However, when tested with users who were not part of the training set, the accuracy fell to around 62\%, suggesting that sign identification from new users is still an open problem.

This dataset provided a good foundation for us to train and test our own model, TSLFormer, for identifying words in Turkish Sign Language.

\begin{figure*}[t]
    \centering
    \includegraphics[width=\textwidth]{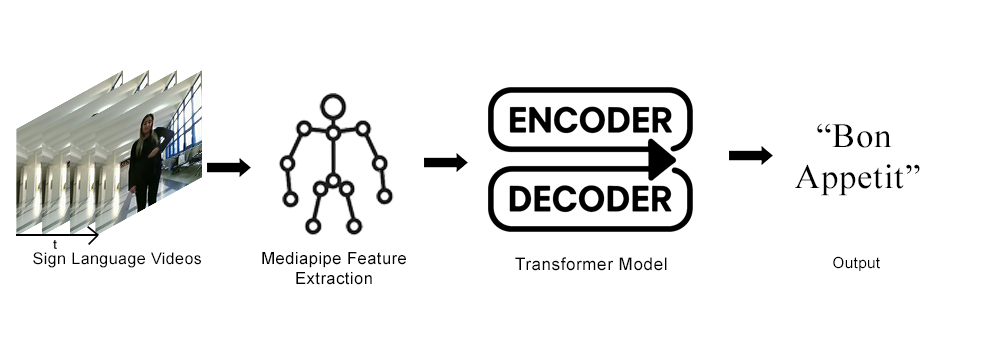}
    \caption{Overall scheme of our proposed TSLFormer-based sign language recognition system.}
    \label{fig:overall}
\end{figure*}

\subsection{Mediapipe Articulation Extraction}

For extracting data related to hands' and upper-body motion dynamics from video recordings, Google's Mediapipe library has been used \cite{lugaresi2019mediapipeframeworkbuildingperception}. Mediapipe is a framework consisting of pre-trained machine learning models specialized in detecting and tracking landmarks or key points on a human's body, including hands and facemarkers. The landmarks are positioned with reference to the major joints like wrists, elbows, fingers, and shoulders. The landmarks are articulated in terms of a three-dimensional coordinate system (X, Y, and Z) and normalized depending on the size of the input image. This approach ensures accurate interpretation of hands and body gestures regardless of orientation and position with respect to the camera.

For our current study, Mediapipe's Pose and Hands modules were utilized for capturing data on articulation of skeletal elements. The Pose module detects major joints on the upper part of the human anatomy, including elbows and shoulders, and the Hands module detects 21 key points on both hands, including fingertip and joints on the tips of the fingers. The coordinates which are extracted enable an overall analysis on movements corresponding to gestures in sign languages. A Python script was created to analyze all videos included in the AUTSL dataset. In the processing process, every video is loaded, and its frames are converted from OpenCV's BGR to RGB format, as required by Mediapipe's functionality. The next step involves the processing of every frame with the Pose and Hands models. When landmarks are successfully detected, X, Y, and Z coordinates on perceivable joints are noted down, otherwise, the corresponding rows are filled with "None" as a placeholder for missing data. The entire aggregated landmark data is placed into an efficiently organized CSV file. Each entry in this file is assigned to a unique video frame and includes video ID, frame index, joint coordinate features, as well as class label for the signed word. This process converts raw visual data into a compact, systematic, and quantifiably informative representation suitable for use with machine learning.

Notably, this method hugely reduces input data dimensionality. A standard 512×512 RGB video frame contains 786,432 pixel-level features, but our landmark-based method reduces this to just 144 features for a frame—made up of 48 keypoints with 3 dimensions (X, Y, Z) each. By excluding redundant visual information and focusing only on motion at the skeletal level, we significantly lower computational load and storage requirements, without sacrificing richness of gesture information.

While dimensionality reduction via landmark extraction greatly reduces computational costs and preserves most dominant semantic gestures, there needs to be some acknowledgment of potential limitations. In removing high-resolution visual details found in complete RGB frames, the system potentially misses out on some motion details such as minor finger articulation, subtle variations of hand shape, or minute wrist rotation. Even though these may seem important, these detail-level features sometimes play to be invaluable for the model to distinguish between visually similar signs. Therefore, although the landmark-based representation yields significant gains in processing efficiency, it comes at a trade-off between computational simplicity and the model's ability to capture extremely detailed gesture information.

The created dataset forms the major input for TSLFormer, our transformer-based model particularly designed for recognition of Turkish Sign Language. The improved representation allows more efficient training while at the same time enriching the robustness and interpretability of the model by focusing on semantically coherent motion features instead of raw pixel intensity.

\begin{figure}[h]
    \centering
    \includegraphics[width=0.4\textwidth]{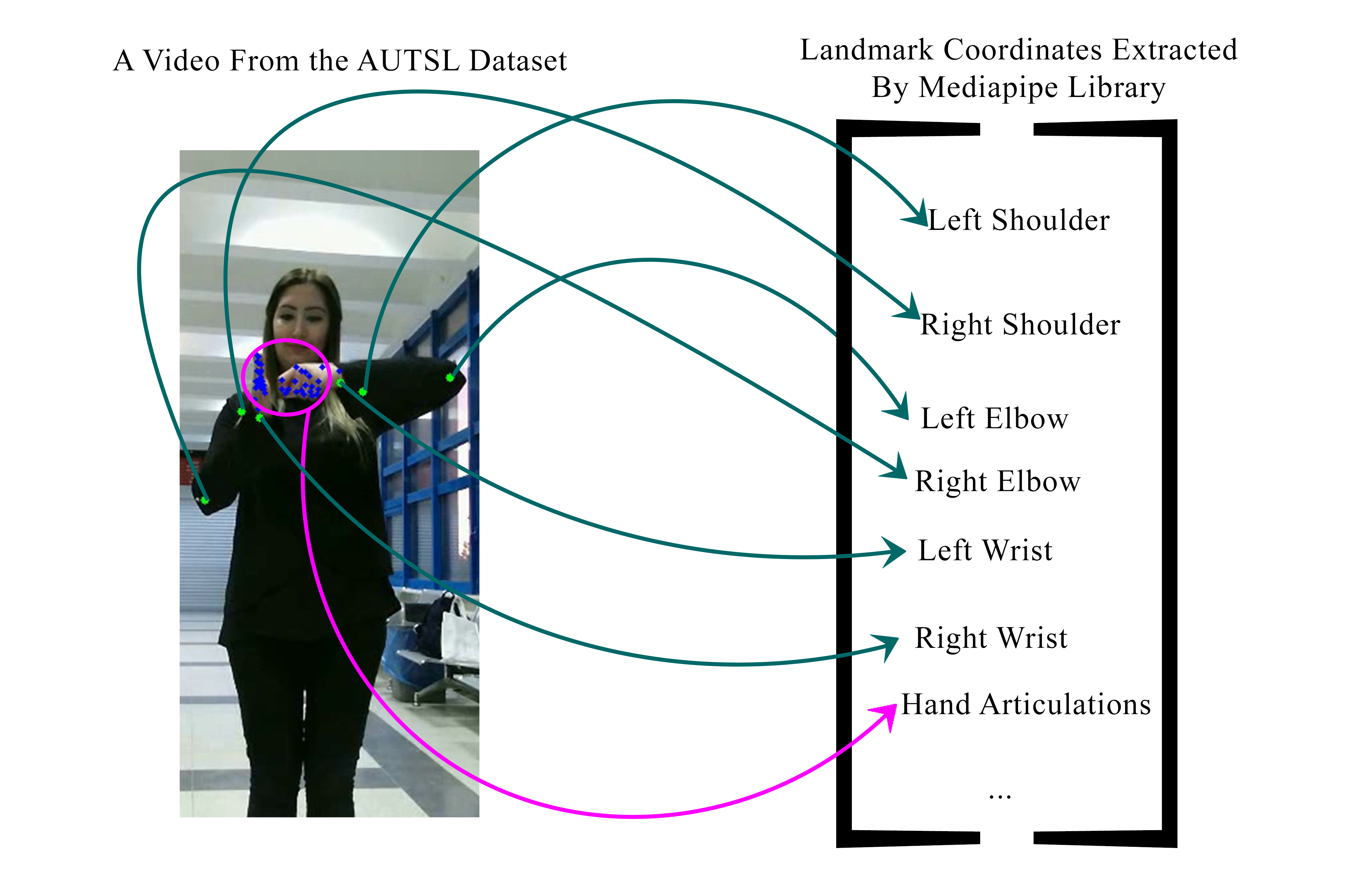}
    \caption{Diagram of Mediapipe Feature Extraction}
    \label{fig:example}
\end{figure}

\subsection{Sign Language Transformer}

Transformers are distinct from the traditional convolutional neural networks (CNNs) and recurrent neural networks (RNNs) since they allow every item in a sequence to attend to every other item directly through self-attention mechanisms \cite{oshea2015introductionconvolutionalneuralnetworks, Rumelhart1986LearningIR}. While CNNs are efficient at learning local spatial patterns and RNNs are efficient at modeling short-term sequential dependencies, they tend to fail at learning long-range temporal relationships and require sequential processing that makes training inefficient. In contrast, transformers calculate entire sequences in parallel and dynamically determine the relative significance of each frame to others, which makes them especially well-suited to modeling the rich, temporally dependent structures of continuous sign language gestures. The ability to model global temporal context over an entire sequence enables more subtle and efficient gesture understanding than local or sequential methods.

Following the conversion of each video into sequences corresponding to the coordinates of body and hand joints using Mediapipe, we proceeded to train a neural network following the Transformer architecture for the purpose of sign classification. Transformers are advanced deep learning models initially developed for natural language processing (NLP) tasks, and they have recently been applied to time-series and video analysis tasks. Since sign language has a sequential and structured nature like spoken language, sign language recognition can be framed as an NLP problem, where the goal is to decipher a sequence of inputs and label them into meaningful units, such as words or phrases. In the current study, we adopt this viewpoint by perceiving the sequence of motion data from each sign video as a type of “sentence” and using a Transformer to perform the analysis. This methodological strategy allows the model to capture temporal patterns and hence improve its competence in distinguishing between signs based on their temporal structure.

Although transformers are widely used in natural language translation, they remain relatively new in the field of sign language recognition, which still primarily relies on convolutional and recurrent neural networks. By treating gesture sequences as linguistic units and applying a transformer-based architecture, TSLFormer aims to offer a fresh perspective on modeling visual language.

Every video was imagined as a sequence of frames, with every frame being described by a feature vector that included the X, Y, and Z locations of specified body and hand landmarks. For each video under analysis, 15 frames per second were chosen to be spaced equally in the course of the video. To manage differences in video lengths and the possibility of missing landmarks in some frames, zero padding was used on the sequences to maintain the uniformity in length for all input sequences, thus making them compatible with batch processing in PyTorch.

The data was structured based on Video ID, with each unique group considered a separate video and any missing values replaced by zeros. Additionally, a special End-of-Sequence (EOS) token was added at the end of each video. The EOS token is another input feature, which helps the model identify the end of the gesture, something important in dealing with sequences of different lengths. 

The model begins with an embedding layer that projects the raw input features into a higher-dimensional representation (e.g., from 144 dimensions to 512 dimensions). Following that, there is a positional encoding layer that adds information about the sequential ordering of the frames to every input vector. Since there are no loops or recurrent structures in RNNs or LSTMs, positional encoding is used to enable understanding of sequence order.

The core building block of the model is a Transformer Encoder, made up of several layers. In each layer, there is a multi-head self-attention mechanism, which enables the model to look at all the frames relative to each other, thus capturing the most important features of the sign \cite{vaswani2023attentionneed}. Then, this is followed by a feedforward network and normalization. To improve the training stability and reduce the likelihood of overfitting, Layer Normalization and Dropout were implemented \cite{ba2016layernormalization, JMLR:v15:srivastava14a}.

Following the passage through the transformer layers, the output from this is then subjected to mean pooling over the entire set of frames, resulting in a combined feature vector that summarizes the entire video. This feature vector is then passed through a final linear (fully connected) layer that makes predictions related to one of the 226 word classes included in the AUTSL dataset.

The model training was done using cross-entropy loss, which is a technique commonly held to be effective in solving classification problems. The Adam optimization algorithm was used, alongside a learning rate scheduler that adjusts the learning rate based on the performance of the model on the validation set\cite{kingma2017adammethodstochasticoptimization}. The performance of the model was measured using accuracy, recall, and F1 score, all of which offered greater insight into the effectiveness of the model, especially in scenarios where there is an imbalanced distribution of the classes.

To make the results more valid, a two-stage validation strategy was employed. First, the dataset was split into an 80\% training set and a 20\% test set, both subsets having a similar distribution of classes. Second, a special kind of K-Fold cross-validation, with K = 4, was applied to the training data, meaning that the model was trained and validated four times, using different portions of the training dataset \cite{CrossValidation}. This approach helped us to minimize the risk of overfitting and offered a more accurate assessment of the model's performance.

At the end of training, we used the test set to evaluate the model's final performance. We also plotted the confusion matrix to visually understand which signs were often confused with others. The backbone of our Turkish Sign Language recognition system is a transformer architecture-based model utilizing landmark-based feature extraction techniques. By providing the model with numerical data extracted from landmarks instead of processing raw video frames directly, the amount of data the model needs to process has been reduced, thereby increasing training efficiency and producing stable classification results.

\begin{figure}[h]
    \centering
    \includegraphics[width=0.4\textwidth]{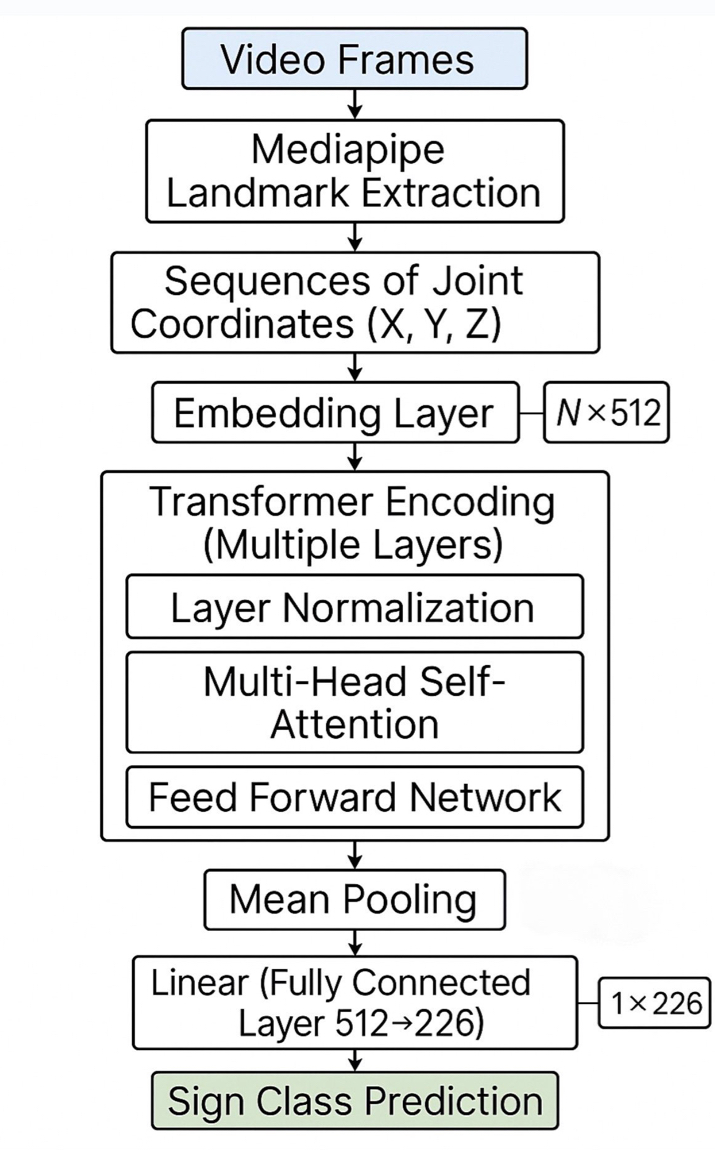}
    \caption{Diagram of TSLFormer}
    \label{fig:example}
\end{figure}

\section{Results}

\subsection{Performance of Test Dataset}

Following the training of TSLFormer, which used features derived from the AUTSL dataset using the Mediapipe framework, we evaluated its performance using 4-Fold Cross Validation in combination with a final test set. The model architecture used hidden layers of 512 dimensions, four attention heads, two encoder layers, and a dropout of 0.2. For stability in the training process, the Adam optimizer was used, working with a learning rate of 1e-4 in combination with a learning rate scheduler \cite{kingma2017adammethodstochasticoptimization}.

The average results obtained from the 4 validation folds are as follows: \begin{table}[h]
\centering
\caption{K-Fold Cross Validation Results}
\begin{tabular}{|c|c|}
\hline
\textbf{Metric} & \textbf{Score (\%)} \\
\hline
Accuracy & 92.85\% \\
Recall   & 92.85\% \\
F1 Score & 93.07\% \\
\hline
\end{tabular}
\label{tab:kfold_results}
\end{table}

After this, we evaluated the model on a separate 20\% test set, which was never seen during training. The final test results were also strong: 

\begin{table}[h]
\centering
\caption{Final Test Results}
\begin{tabular}{|c|c|}
\hline
\textbf{Metric} & \textbf{Score (\%)} \\
\hline
Accuracy & 90.67\% \\
Recall   & 90.67\% \\
F1 Score & 90.67\% \\
\hline
\end{tabular}
\label{tab:kfold_results}
\end{table}

The results show that our approach, based on transformer architecture and augmented by the joint landmark data received via Mediapipe, has great potential for correct Turkish Sign Language word recognition. Additionally, the sole use of 3D joint coordinates, rather than full RGB or depth video data, improves the efficiency of the system and makes it suitable for real-time use.

\begin{figure}[ht]
    \centering
    \includegraphics[width=0.4\textwidth]{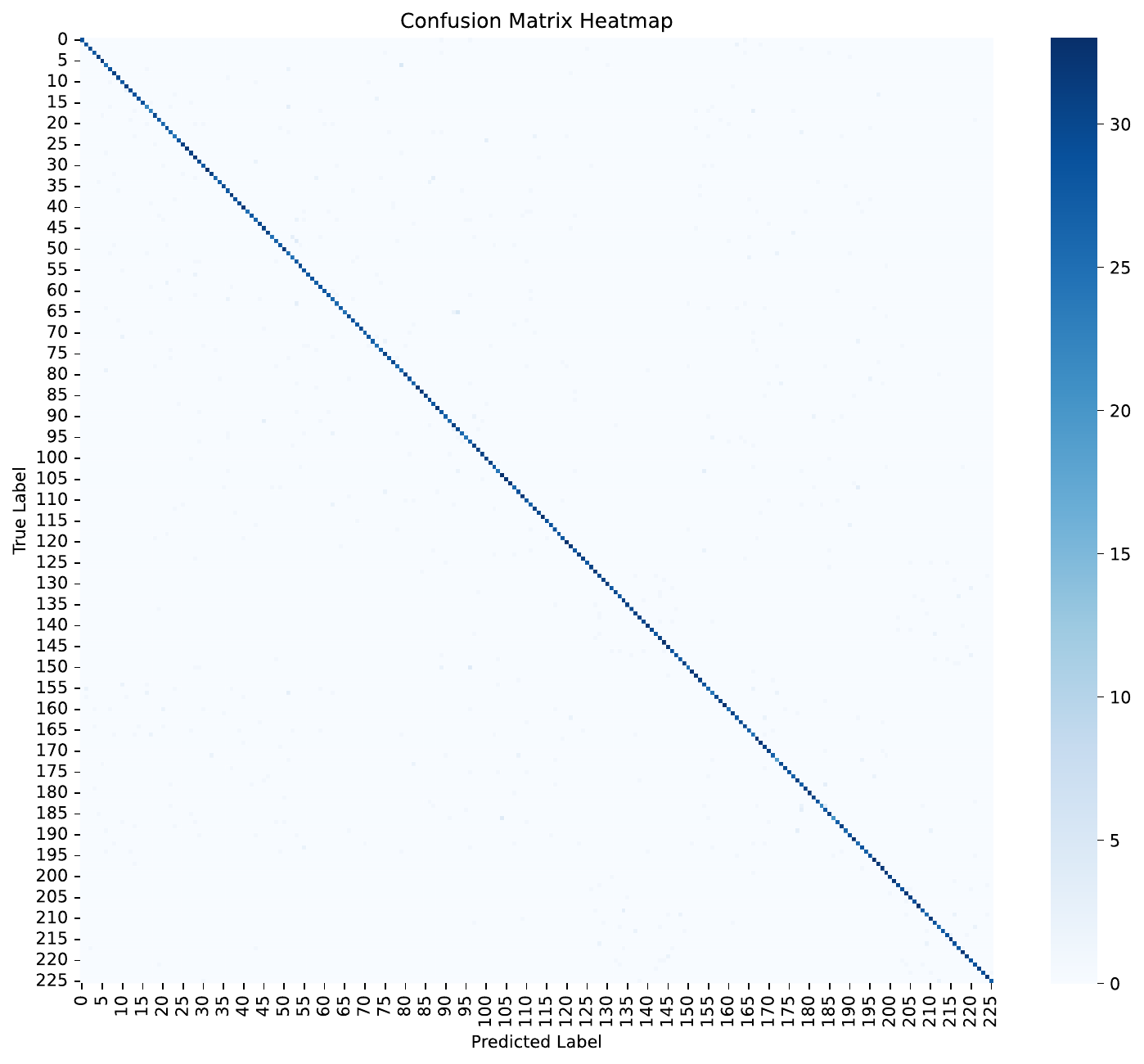}
    \caption{Confusion Matrix for 226 Classes and TSLFormer}
    \label{fig:confmatrix}
\end{figure}

In addition to overall evaluation metrics, we also analyzed the confusion matrix to understand how well the model performs across all 226 sign classes. As shown in Figure \ref{fig:confmatrix}, the confusion matrix is largely diagonal, indicating that the model makes correct predictions for most signs. The high density along the diagonal line suggests that the model is consistently assigning the correct label for each class. Very few off-diagonal entries appear, which means that misclassifications are rare and typically occur only between visually or motion-wise similar signs. This further supports that the model has learned to distinguish between a wide range of Turkish Sign Language words using only skeletal joint information.

To compare the performance of TSLFormer with other methods on the same dataset, we did an extensive survey of various relevant studies that have used the AUTSL dataset. Their results are summarized in \ref{tab:autsl_comparison}.

\begin{table*}[htbp]
\centering
\caption{Comparison with Existing Work on the AUTSL Dataset}
\resizebox{\textwidth}{!}{%
\begin{tabular}{|l|c|l|}
\hline
\textbf{Method} & \textbf{Accuracy (\%)} & \textbf{Used Modalities} \\
\hline
\textbf{Our Method (Transformer + Mediapipe)} & \textbf{90.67} & Hand joints (Mediapipe) \\
SAM-SLR \cite{jiang2021skeletonawaremultimodalsign} & 98.42 & RGB + Skeleton + Face \\
S3D \cite{9523152} & 98.34 & RGB + Skeleton + Face \\
TD-SL \cite{10222729} & 97.93 & RGB + Top-Down Hand Attention \\
USTC-SLR \cite{ustc2021chalearn} & 97.62 & RGB + Skeleton + Face \\
Jalba \cite{jalba2021fact} & 96.15 & RGB + Skeleton + Face \\
VLE-Transformer \cite{gruber2021mutualsupport} & 95.46 & RGB + Skeleton + Face \\
RGB-MHI \cite{Mercanoglu_Sincan_2022} & 93.53 & RGB + Motion History Images \\
VTN-PF \cite{VTN-PF} & 92.92 & RGB + Pose Flow \\
Baseline \cite{Sincan_2020} & 49.22 & RGB \\
\hline
\end{tabular}
}
\label{tab:autsl_comparison}
\end{table*}

These results show that while some models achieve higher accuracy, they often rely on complex architectures and multiple input types such as RGB video, depth information, skeleton data, and facial features. In contrast, TSLFormer reaches nearly 90\% accuracy using only Mediapipe landmark data, which makes it much simpler and more efficient to implement, especially in systems where computational resources are limited.

This comparison confirms that our proposed method is a valid and effective alternative for Turkish Sign Language recognition, especially in lightweight or real-time applications.

\subsection{Demo Application}

For demonstration purposes, we created an application implemented with Python that records sign gestures through a webcam and performs real-time inferences on a single-word basis with a pre-trained model. The system uses live video input, landmark detection provided by Mediapipe, and sequence classification provided by the TSLFormer model.

In the demo, the system continuously captures webcam input and starts recording when motion is detected. When a sign gesture is completed—a period of little movement is detected—the system will automatically stop recording and start processing the recorded frames. Mediapipe's Pose and Hands modules are called to extract landmarks representing upper body and hand joints from each frame \cite{lugaresi2019mediapipeframeworkbuildingperception}. 15 frames per second are evenly sampled from the recorded video, and each frame is converted to a feature vector made up of 3D coordinates. The features are cleaned, normalized, and padded when necessary, then passed to TSLFormer for classification.

After the inference stage, the system produces the five most likely lexical categories and their respective probabilities. This method not only provides the most likely choice but also produces a hierarchically structured list of alternatives, thus providing insight into the distribution of certainty within the model.

During the evaluation process, TSLFormer showed reliability in identifying a wide range of indicators. Oftentimes, it assigns high probability to its most preferred prediction, even in cases of misprediction. While such a pattern is expected when the model makes correct predictions, it was observed that TSLFormer is very certain about its chosen alternative when predicting incorrectly. Such a finding emphasizes the need for better calibration or methods for uncertainty estimation, allowing the model to better express uncertainty. There is a perceivable trend in errors in prediction where TSLFormer commonly misidentifies signs appearing at both hands' convergence. The misinterpretation is likely due to the limited resolution of skeletal joint information, making it difficult to differentiate slight differences in closeness of hands and motion trajectories amid concurrent occurrences. These signs tend to have similar features in terms of landmark coordinates and motion vectors, particularly when detailed information at the level of the fingers is not captured satisfactorily.

Despite these challenges, the demonstration system is able to correctly identify several isolated signs with significant accuracy and low latency, thus illustrating the potential value of TSLFormer for real-time sign recognition applications. In addition, these results support the main contribution of this work: that skeletal landmark information, separate from full RGB video, can be used to create efficient, practical, and effective sign recognition systems.

\section{Discussion}

In this study, we presented an exclusive Turkish Sign Language (TİD) recognition system based on the use of skeletal joint coordinates derived using the Mediapipe framework. As opposed to attempting the more complex task of identifying letters based on the temporal dynamics of the movement involved in signs, this method focused on identifying words at the level of the gesture. The system is designed for comparing coordinate-based features derived for 16 anchor frames for each of the videos and classifying these features via a transformer-inspired neural network.

The results show that TSLFormer performs admirably well even using only pose and hand landmarks as input features. With an average validation accuracy of 92.85\% and a final test accuracy of 90.67\%, the model confirms that a simplified gesture representation—extracted only from joint data—can be sufficient to enable high-accuracy classification. This suggests that the approach is especially beneficial for real-time settings or low-resource settings, where the processing of RGB video and depth data might not only be computationally expensive but also unnecessary \cite{Berkan}.

A comparative study with the literature that uses the same AUTSL dataset reveals that several top-performing models use complex and multi-modal inputs. For example, SAM-SLR as well as S3D achieves accuracy scores in excess of 98\% using RGB imagery, skeletal data, as well as facial features, often making use of ensemble or multi-stream approaches \cite{jiang2021skeletonawaremultimodalsign, 9523152}. Similarly, TD-SLR achieves accuracy at the level of 97.93\% using top-down attention mechanisms as well as optical flow information together with RGB data. The operational requirements for such models include long training times, higher memory capacity, as well as higher hardware requirements, possibly making them unsuitable for use in real-world scenarios. Conversely, our approach only uses the joint coordinates of hands and upper body from Mediapipe. By not needing RGB video, depth maps, or facial features, our system is able to reach an accuracy level of nearly 90\%, while being significantly lighter and easier to implement. This makes our approach a viable choice for a number of applications, from embedded systems, mobile, or browser-based applications, to help improve the accessibility of those in the hearing-impaired community.

Confusion matrix is used as the validation for the effectiveness of the model \cite{CrossValidation}. Most signs were appropriately classified, with few misclassifications happening among signs that share visual or movement similarities. This suggests that the model has successfully learned important spatial-temporal features in the joint data, thus allowing it to clearly discriminate between a vast vocabulary in Turkish Sign Language.

The results of our experiment suggest that accurate recognition of sign language is possible with the use of Mediapipe landmark information combined with the transformer-based framework \cite{vaswani2023attentionneed}. 

While TSLFormer is behind the best-performing models in terms of raw accuracy, the construction decisions made in its design, and the working principles of the model make it particularly attractive for real-world deployment. Unlike approaches requiring RGB, depth, or multi-modal input streams, TSLFormer operates using only lightweight skeletal joint data, significantly reducing computational demands. This allows for faster inference, reduced memory needs, and easier integration into edge devices, e.g., mobile devices or assistive equipment. As a result, the model's trade-off of slightly lower accuracy for substantially higher efficiency renders it a viable and scalable option for real-time Turkish Sign Language recognition systems.

\section{Conclusion and Future Work}

In this study, we developed a Turkish Sign Language recognition system using a transformer-based model trained on features obtained from video frames through Google's Mediapipe library \cite{lugaresi2019mediapipeframeworkbuildingperception}. Instead of using full RGB videos, our approach focused only on the 3D coordinates of hand and upper body landmarks, which caused a great reduction in data size and computational requirements. We trained and tested TSLFormer on the AUTSL dataset, which includes over 36,000 videos for 226 unique words \cite{Sincan_2020}. TSLFormer achieved a final test accuracy of 90.67\%, which shows that sign language recognition can be successfully carried out using only joint-based features.

As opposed to other methodologies reported in the literature that make use of complicated structures and multiple input modalities such as RGB, depth, and facial features, our solution is significantly simpler but just as accurate. This makes TSLFormer very suitable for use in real-time environments or on mobile devices, where limitations on resources available can be a determining factor.

Future work could elaborate on our system for facilitating sentence-level recognition, hence enabling detection of sequential signs as opposed to mere isolated lexical detection. In addition, the creation of a real-time version based on webcam technology would enhance its interactive capacity considerably. Moreover, evaluation of the model in multiple sign language datasets or in combined datasets may shed light on its transferability for novel signers as well as for different recording setups. Finally, integrating domain adaptation methods or optimization of the model for multilingual signs would enhance the system's inclusiveness as well as overall usability.


\end{document}